\pdfoutput=1

\documentclass[11pt]{article}

\usepackage{acl}

\usepackage{times}
\usepackage{latexsym}
\usepackage{microtype}
\usepackage{graphicx}
\usepackage{adjustbox}
\usepackage{multirow}
\usepackage{booktabs}
\usepackage{tabularx}
\usepackage[inline]{enumitem}
\usepackage{flushend}
\usepackage{balance}
\usepackage[T1]{fontenc}

\usepackage[utf8]{inputenc}

\usepackage[bottom]{footmisc}
\usepackage[section]{placeins}

\newcommand{\round}{\operatorname{round}}
\newcommand{\abs}{\operatorname{abs}}
\newcommand{\erf}{\operatorname{erf}}

\usepackage{amsmath}
\usepackage{csquotes}

\title{Empirical Evaluation of Post-Training Quantization Methods for Language Tasks}

\author{Ting Hu \\
  Hasso Plattner Institute\\
  University of Potsdam \\
  Potsdam, Germany \\
  \texttt{ting.hu@hpi.de} \\
  \And
  Christoph Meinel\\
  Hasso Plattner Institute\\
  University of Potsdam \\
  Potsdam, Germany \\
  \texttt{meinel@hpi.de}\\
  \And
  Haojin Yang \\
  Hasso Plattner Institute\\
  University of Potsdam \\
  Potsdam, Germany \\
  \texttt{haojin.yang@hpi.de} \\
  }

\begin{document}
\maketitle

\begin{abstract}
Transformer-based architectures like BERT have achieved great success in a wide range of Natural Language tasks. Despite their decent performance, the models still have numerous parameters and high computational complexity, impeding their deployment in resource-constrained environments. Post-Training Quantization (PTQ), which enables low-bit computations without extra training, could be a promising tool. In this work, we conduct an empirical evaluation of three PTQ methods on BERT-Base and BERT-Large: Linear Quantization (LQ), Analytical Clipping for Integer Quantization (ACIQ), and Outlier Channel Splitting (OCS). OCS theoretically surpasses the others in minimizing the Mean Square quantization Error and avoiding distorting the weights' outliers. That is consistent with the evaluation results of most language tasks of GLUE benchmark and a reading comprehension task, SQuAD. Moreover, low-bit quantized BERT models could outperform the corresponding 32-bit baselines on several small language tasks, which we attribute to the alleviation of over-parameterization. We further explore the limit of quantization bit and show that OCS could quantize BERT-Base and BERT-Large to 3-bits and retain 98\% and 96\% of the performance on the GLUE benchmark accordingly. Moreover, we conduct quantization on the whole BERT family, i.e., BERT models in different configurations, and comprehensively evaluate their performance on the GLUE benchmark and SQuAD, hoping to provide valuable guidelines for their deployment in various computation environments.
\end{abstract}

\section{Introduction}

The emergence of large pre-trained language models like BERT \cite{kenton2019bert} has facilitated the development of the Natural Language Processing (NLP) area with numerous real-life applications. The mainstream paradigm in NLP is pre-training + fine-tuning, i.e., firstly pre-training language models with massive amounts of unlabeled texts and then fine-tuning on various downstream tasks, such as sentence sentiment classification and reading comprehension. However, pre-trained models \cite{liu2019roberta,yang2019xlnet,fedus2021switch} have an increasing number of parameters and are computationally expensive even for fine-tuning, limiting their further deployment on limited-resource devices. Moreover, large language models could be over-parameterized on small downstream tasks. These attract much research attention on BERT model compression, aiming to decrease the model size, accelerate inference, and reduce energy consumption.

There are several strategies of model compression involving Quantization, Knowledge Distillation, Pruning, and Parameter Sharing. We focus on model quantization, performing computations with lower-bit than 32-bit floating-point precision. Quantization-Aware Training (QAT) and Post-Training Quantization (PTQ) are two dominant approaches. QAT requires an extra fine-tuning process to adjust the quantized parameters using training data, whereas PTQ does not. Most PTQ related works \cite{kim2021bert,dai2021vs,shen2020q} quantize both the weights and the activations of BERT models. In contrast, we merely quantize the weights, considering quantizing activations incurs extra real-time computational overhead or beforehand calculation using a series of calibration inputs \cite{gholami2021survey}. Additionally, we quantize the weights to the same bit rather than different bits, which is easy to use and hardware-friendly. We empirically evaluate three post-training quantization methods: Linear Quantization (LQ), Analytical Clipping for Integer Quantization (ACIQ) \cite{ron2018aciq}, and Outlier Channel Splitting (OCS) \cite{zhao2019improving}  on the GLUE benchmark \cite{wang2018glue} and a reading comprehension task, SQuAD \cite{rajpurkar2016squad}.

Our experimental results show that applying PTQ methods to quantize the weights effectively reduces the memory footprint of BERT models with minor and acceptable performance degradation. Furthermore, we provide a comprehensive evaluation of different BERT configurations in this work in the hope of offering helpful guidance for their practical deployment. We highlight some interesting findings regarding BERT-Base and BERT-Large below.
\begin{itemize}
\item 8-bit BERT-Base quantized by ACIQ or OCS outperforms the 32-bit baseline on the average score of the GLUE benchmark.
\item We can quantize BERT-Base and BERT-Large to 3-bit by OCS and retain 98\% and 96\% of the performance on GLUE benchmark, respectively.
\item Though 32-bit BERT-Large consistently performs better than 32-bit BERT-Base, quantized BERT-Base could surpass quantized BERT-Large taking both the model size and the performance into consideration.
\end{itemize}

\section{Related work}

The aim of model compression is obtaining models in smaller size, less computation, and faster inference without severe performance deterioration. There are different approaches to achieve these: model pruning \cite{fan2019reducing,michel2019sixteen,hou2020dynabert}, Knowledge Distillation \cite{sun2019patient,sanh2019distilbert,jiao2019tinybert}, parameter sharing \cite{dehghani2018universal,lan2019albert}, and model quantization \cite{zafrir2019q8bert,zhang2020ternarybert,bai2020binarybert,zadeh2020gobo,kim2021bert,dai2021vs,shen2020q}.

Pruning is beneficial to memory and bandwidth reduction by removing redundant parameters or neurons. Study on BERT model pruning involves a variety of granularities. \citet{fan2019reducing} randomly drops BERT layers during training and prune shallow sub-networks on-demand at inference time. \citet{michel2019sixteen} observes that Multi-Head Attention is not always superior, and a large portion of attention heads could be pruned during inference. \citet{hou2020dynabert} shows that both the BERT layers and the attention heads in each layer could be selectively pruned with reasonable performance reduction or sometimes even a slight performance improvement.

Knowledge Distillation (KD) aims to transfer the knowledge of a large teacher model to a smaller student model. The student generally imitates the behaviors of the teacher model, for instance, the final outputs and intermediate hidden states. Current works conduct KD in different stages. \citet{sun2019patient} encourages a 6-layer student model to patiently learn from multiple intermediate layers of BERT-Base on downstream tasks. \citet{sanh2019distilbert} obtains a general-purpose 6-layer student model via KD during the pre-training phase. \citet{jiao2019tinybert} further builds a 4-layer model TinyBERT and conducts KD during both the pre-training and the fine-tuning stage, which is time-consuming and energy-intensive.

Parameter sharing is generally used to improve parameter efficiency by reusing parameters among separate components of neural networks. \citet{dehghani2018universal} shares weights across the sequential computation steps of Transformer \cite{vaswani2017attention}, resulting in a parallel-in-time recurrent sequential model. \citet{lan2019albert} proposes all parameters sharing across BERT layers and designs A Lite BERT (ALBERT) architecture with fewer parameters and faster training speed than traditional ones.

We categorize model quantization into Quantization-Aware Training (QAT) and Post-Training Quantization (PTQ). In terms of QAT, Q8BERT \cite{zafrir2019q8bert} quantizes the weights and activations of BERT to 8-bits by introducing quantization error during the fine-tuning stage. QBERT \cite{shen2020q} maintains the 8-bit activations and conduct mixed-precision quantization on the weights to minimize performance degradation, which could be unfriendly to some hardwares. TernaryBERT \cite{zhang2020ternarybert} uses 3-bit weights and 8-bit activations and manages to overcome performance degradation by a KD method.
BinaryBERT \cite{bai2020binarybert} binarizes the weights and quantizes the activations to 4-bit by further refinement on TernaryBERT using augmented data, which is time- and resource-consuming. BiBERT \cite{qin2022bibert} is a fully binary BERT while the performance on some data sets, for example, the CoLA and STS-B, are still unacceptable.
PTQ generally does not require extra fine-tuning or retraining to make up for the quantization error. GOBO \cite{zadeh2020gobo} quantizes BERT to 3-bit precision by handling the vast majority of weights that comply with the Gaussian and the few outliers separately, while the outliers are still saved as 32-bit. I-BERT \cite{kim2021bert} achieves INT8 inference based on specialized integer-only approximation methods for nonlinear operations. \citet{dai2021vs} quantizes both weights and activations to 4-bit by per-vector quantization while a calibration process is needed for the scaling factors.

Our work falls into the PTQ category and differentiates from the others in the following aspects. Firstly, we only quantize the weights rather than both weights and activations. Quantizing the activations requires
a calibration process beforehand to obtain the optimal activation parameters and keep this information for inference\cite{dai2021vs}. This incurs some extra computation while ours does not. Secondly, unlike GOBO \cite{zadeh2020gobo}, we quantize the weights to the same bit, which is easy to use and hard-ware friendly \cite{zhang2020ternarybert}. Thirdly, our work offers a comprehensive evaluation of the quantized BERT family, ranging from BERT-Tiny to BERT-Large, by applying the quantization method discussed in the next section.

%quantize weights + activations: I-BERT\cite{kim2021bert}, VS-Quant\cite{dai2021vs}(per-vector scaled quantization), QBERT\cite{shen2020q}(group-wise quantization).
%quantize weights: Q8BERT\cite{zafrir2019q8bert}(quantization-aware training, symmetric linear quantization), GOBO\cite{zadeh2020gobo}(a dictionary to index outliers, 3bits).
%quantization + kd: tenary-BERT\cite{zhang2020ternarybert}, quantization-aware and distillation-aware.
%binary stuff: binaryBERT\cite{bai2020binarybert}: quantization-aware training + BiBERT.

\section{Methodology}

\subsection{BERT architecture}

\begin{figure}[!th]
\centering
\includegraphics[width=0.8\linewidth]{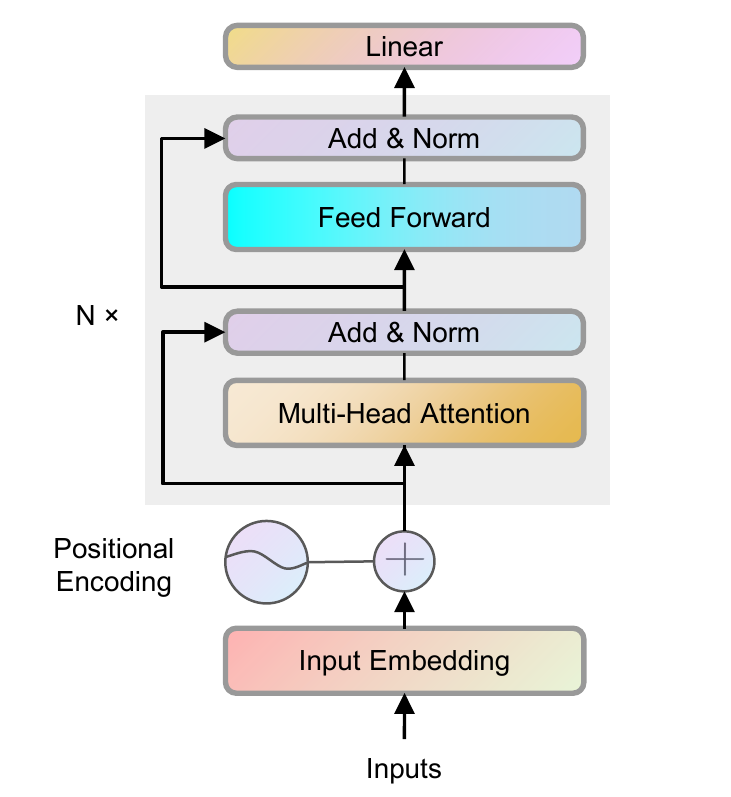}
\caption{BERT architecture.} \label{bert_arc}
\end{figure}

The architecture of BERT \cite{devlin2018bert} is depicted in Fig.~\ref{bert_arc}, consisting of the embedding layer and a stack of $N$ intermediate layers. Each layer comprises a Multi-Head Attention module and a fully connected Feed-Forward Network. The model is pre-trained with large amounts of unlabeled text data by the Masked Language Modeling objective. When fine-tuning, an additional linear layer is appended and optimized together with BERT on each downstream Natural Language Understanding task. The most commonly used BERT configurations are BERT-Base and BERT-Large, of which the detailed architecture and size of different components are shown in the Tab.~\ref{bert_arc}.
% (24 layers, 16 attention heads, and 340M parameters) (12 layers, 12 attention heads, and 110M parameters) The model size of 32-bit BERT-base and BERT-large are 436MB and 1.3GB, respectively.

%\begin{table}[!th]
%\begin{center}
%    \begin{tabular}{l|cc}
%    \hline
%    \multicolumn{3}{c}{Model Configuration} \\
%    \hline
%    & BERT-Base & BERT-Large \\
%    \hline
%    \# Layers & 12 & 24 \\
%    \# Attention Heads & 12 & 16 \\
%    \# Hidden Size & 768 & 1024 \\
%    \# Intermediate Size & 3072 & 4096 \\
%    \# Parameters & 110M & 340M \\
%    \hline
%    \multicolumn{3}{c}{Model Size} \\
%    \hline
%    Embedding Layer & 89 MB & 119 MB \\
%    Weights & 326 MB & 1.12 GB \\
%    \hline
%    \end{tabular}
%\end{center}
%\caption{\label{bert_arc} The configuration and model size of BERT-Base and BERT-Large. }
%\end{table}

\subsection{Quantization}

Linear Quantization (LQ) maps the inputs to a set of discrete, evenly-spaced grid values, and its scheme can be formulated as Eq.~\ref{lq}, where $X_{f}$ is an $n$-bit floating-point weight tensor, $s$ is the scaling factor, $z$ is the zero point, $X_{q}$ is the corresponding quantized tensor, and $k$ is the quantization bit. We simplify the discussion to symmetric Linear Quantization, where the zero point $z$ is zero and the scaling factor $s$ is computed as Eq.~\ref{slq}.
\begin{equation}
\label{lq}
\begin{split}
  & X_{q}=\round ( \frac{X_{f}}{s} +z) \\
  & s = \frac{2^{k}-1}{\max(X_{f})-\min(X_{f})} \\
  & z = \frac{\min(X_{f})\times (1-n)}{\max(X_{f})-\min(X_{f})}
\end{split}
\end{equation}

\begin{equation}
\label{slq}
  s=\frac{2^{k}-1}{\max(\abs(X_{f}))}
\end{equation}

Dequantization transforms the quantized tensor back to a floating-point tensor $\hat{X_{f}}$ in Eq.~\ref{dq}, given the scaling factor $s$.
\begin{equation}
\label{dq}
 \hat{X_{f}}=X_{q} \times s
\end{equation}
The Mean Square quantization Error (MSE) between the input tensor $X_{f}$ and its dequantized counterpart $\hat{X_{f}}$ is defined in Eq.~\ref{mse}. Obviously, MSE is largely influenced by the scaling factor $s$, which is sensitive to the largest input $\max(\abs(X_{f}))$, i.e., the outliers.
\begin{equation}
\label{mse}
 MSE=E[(X_{f}-\hat{X_{f}})^{2}]
\end{equation}

In order to reduce the MSE, some works \cite{ron2018aciq, zadeh2020gobo} limit the input tensor to a certain range by clipping the outliers. In this case, the clipping threshold is significant. ACIQ \cite{ron2018aciq} employs a closed-form solution for the clip threshold to minimize the MSE, assuming the weight's distribution is either a Gaussian or a Laplacian.
According to the analysis of \cite{zadeh2020gobo}, the vast majority of weights per layer closely follow some Gaussian distribution with very few outliers, which makes ACIQ appropriate for BERT quantization here. The optimal clipping value $\alpha$ minimizing the MSE is obtained by setting the derivative in Eq.~\ref{aciq} equal to zero, where $\sigma$ is the standard deviation of the weight distribution, and $k$ is the quantization bit. This method is convenient and fast without candidate threshold sweeping. We take the open source code and modify from this link\footnote{\url{https://github.com/submission2019/AnalyticalScaleForIntegerQuantization}}.
\begin{equation}
\label{aciq}
\begin{split}
 \frac{\partial E[(X_{f}-\hat{X_{f}}]^{2}}{\partial \alpha}=\alpha[1-\erf(\frac{\alpha}{\sqrt{2} \sigma})]-\frac{\sigma ^{2}e^{-\frac{\alpha ^2}{2\sigma ^2}}}{\sqrt{2\pi} \sigma}\\-\frac{\sigma e^{-\frac{\alpha ^2}{2\sigma ^2}}}{\sqrt{2\pi}} + \frac{2\alpha}{3 \cdot 2^{2k}}
\end{split}
\end{equation}

Though clipping reduces the overall MSE by narrowing the range of the weight distribution, the outliers potentially indicating important information are greatly distorted. \citet{zhao2019improving} proposes Outlier Channel Splitting (OCS) to avoid this issue by splitting the outliers and moving them toward the center of the distribution. The approach first duplicates the neurons, then halves their outputs or halve the outgoing weight connections. Consider a feed-forward linear layer in BERT, which can be defined as $Y=XW$, where $X \in R^{l\times h}$, $Y\in R^{l\times d}$, $W\in R^{h\times d}$ refer to the input tensor, the output tensor, and the weight, respectively. Moreover, $l$ is the sequence length, $h$ is the hidden size, and $d$ is the intermediate feed-forward size. We duplicate all the input neurons of $W$ and halve the weights. Then the same outputs are maintained by Eq.~\ref{ocs}. Nevertheless, the maximum MSE is doubled by this naive OCS, since the halves $W/2$ may be rounded in the same direction. \citet{zhao2019improving} then proves that quantization-aware splitting in Eq.~\ref{ocs1} is optimal. We use the code from the authors\footnote{\url{https://github.com/cornell-zhang/dnn-quant-ocs}}.
\begin{equation}
\label{ocs}
 Y=
 \begin{bmatrix}
 X & X
\end{bmatrix}
\begin{bmatrix}
 W/2 \\
 W/2
\end{bmatrix}
\end{equation}

\begin{equation}
\label{ocs1}
 Y=
 \begin{bmatrix}
 X & X
\end{bmatrix}
\begin{bmatrix}
 (W-0.5)/2 \\
 (W+0.5)/2
\end{bmatrix}
\end{equation}
This makes the model work equivalently at the cost of model size overhead since the number of input channels is doubled. In practice, only the channels containing the largest absolute values in the layer are subject to splitting. An expansion ratio $r$ is set in this method, determining the approximate level of acceptable overhead in the network. We set it to be 0.01 in our implementation.

In this work, we employ three quantization methods aforementioned: LQ, ACIQ, and OCS, and quantize the embedding layer and all the linear sub-networks in each BERT layer except for the final classifier, which is sensitive to quantization and vital for predictions. The quantization ratio is 99.6\%, calculated by the number of quantized weights divided by the total number of weights. The performance of these quantized models is elaborated in the following section.

\section{Experiments}

\subsection{Downstream language tasks and evaluation metrics}

We evaluate low-bit BERT models on General Language Understanding Evaluation (GLUE) benchmark \cite{wang2018glue} and a reading comprehension task SQuAD v1.1 \cite{rajpurkar2016squad}. GLUE consists of several Natural Language Understanding tasks: single sentence tasks: CoLA \cite{warstadt2019neural} and SST-2 \cite{socher2013recursive}, similarity and paraphrase tasks: MRPC \cite{dolan2005automatically}, STS-B \cite{cer2017semeval}, and QQP \cite{chen2018quora}, inference tasks: MNLI \cite{williams2017broad}, QNLI \cite{rajpurkar2016squad}, and RTE \cite{bentivogli2009fifth}. We exclude WNLI \cite{levesque2012winograd} since this is a relatively small dataset and shows an unstable behaviour.

In terms of evaluation metrics, we report Matthews Correlation for CoLA, Accuracy for QNLI, RTE, and SST-2, the average of Accuracy and F1 for MRPC and QQP, the average of Pearson Correlation and Spearman Correlation for STS-B. The average Accuracy of MNLI-match (MNLI-m) and MNLI-mismatch (MNLI-mm) is recorded as the Accuracy of MNLI. With regard to SQuAD, we report Exact Match (EM) and F1.

\begin{table*}[!th]
\small
\begin{center}
    \begin{tabular}{c|ccccccccc}
    \hline
    \multicolumn{9}{c}{BERT-Base} \\
    \hline
     & MRPC & SST-2 & STS-B & CoLA & RTE & QNLI & MNLI  & QQP & Avg\\
    \hline
    32-bit & 86.03 & \textbf{92.09} & 89.05 & 56.50 & 65.34 & 90.61 & 84.33 & \textbf{88.91} & 81.61 \\
    LQ & \textbf{86.52} & 91.28 & 89.04 & 54.02 & 64.98 & 90.66 & 84.20 & 88.20 & 81.11(-0.50)\\
    ACIQ & 86.28 & 91.51 & 89.07 & \textbf{58.33} & \textbf{65.70} & \textbf{90.72} & 84,15 & 88.28 & \textbf{81.76(+0.15)} \\
    OCS & 86.42 & 91.97 & 89.07 & 57.27 & \textbf{65.70} & \textbf{90.72} & \textbf{84.39} & 88.44 & 81.75(+0.14) \\
    \hline
    \multicolumn{9}{c}{BERT-Large} \\
    \hline
     & MRPC & SST-2 & STS-B & CoLA & RTE & QNLI & MNLI & QQP & Avg \\
    \hline
    32-bit & 89.13 & 93.35 & \textbf{89.39} & \textbf{62.84} & 70.40 & \textbf{92.62} & \textbf{91.00} & \textbf{89.42} & \textbf{84.77} \\
    LQ & 88.10 & 93.23 & 89.37 & 61.85 & 69.31 & 92.55 & 86.53 & \textbf{89.42} & 83.80(-0.97) \\
    ACIQ & 88.33 & 93.23 & 89.09 & 61.83 & 70.04 & 92.39 & 86.30 & 89.26 & 83.81(-0.96) \\
    OCS & \textbf{89.30} & \textbf{93.80} & 89.32 & 62.33 & \textbf{71.11} & 92.53 & 86.61 & 89.31 & 84.29(-0.48) \\
    \hline
    \end{tabular}
\end{center}
\caption{\label{glue_benchmark} We quantize 32-bit baseline models to 8-bits by three quantization methods. \enquote{Avg} represents the average score of all tasks. We report the score difference between the baseline and the quantized model in parentheses. ACIQ and OCS bring about a slight performance improvement on BERT-Base. }
\end{table*}

\begin{figure*}[!th]
\centering
\includegraphics[width=0.7\linewidth]{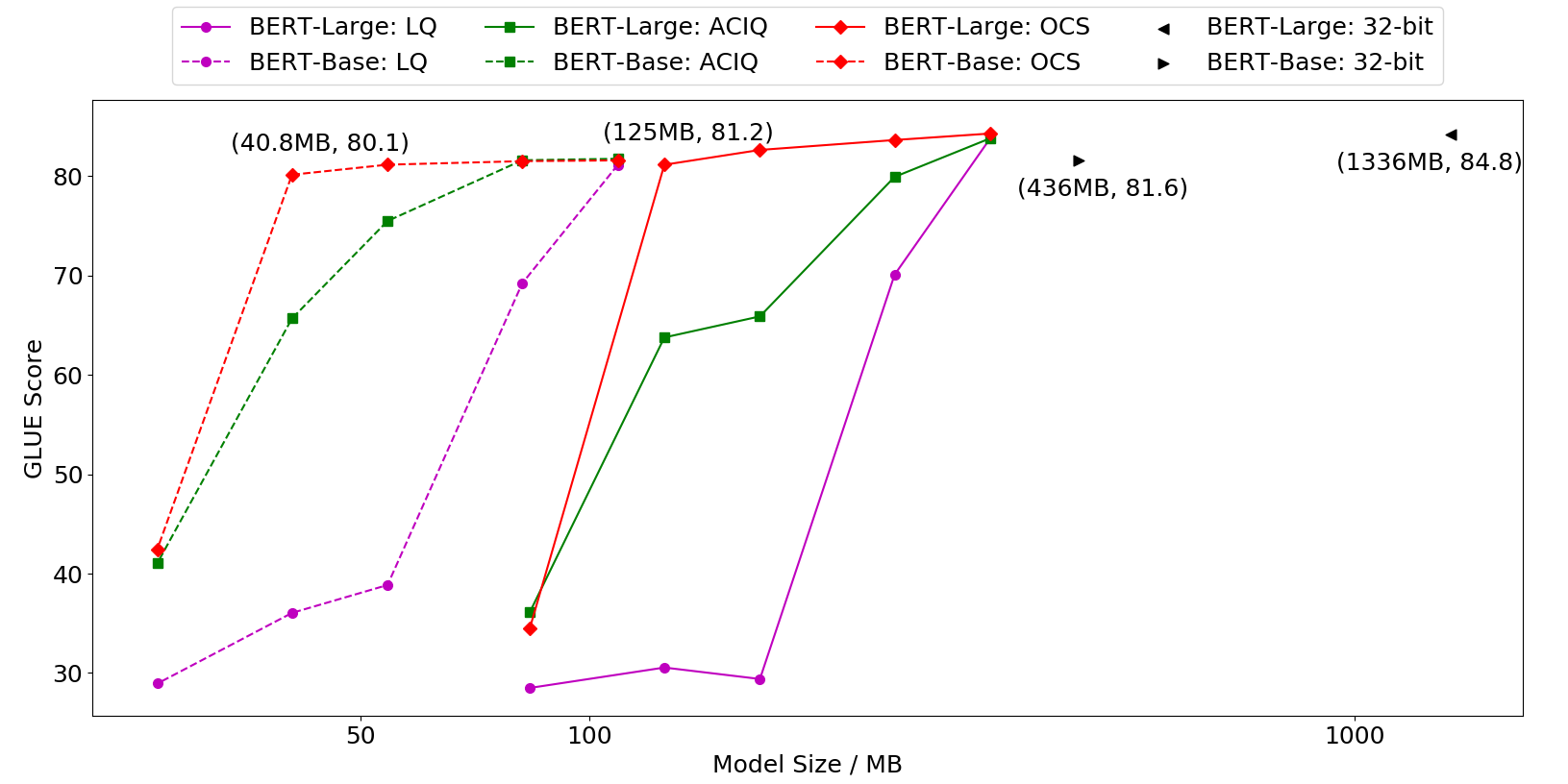}
\caption{The performance of quantized BERT-Base and BERT-Large on the dev set of GLUE benchmark. Three quantization methods, LQ, ACIQ, and OCS, are applied. The points shown from right to left correspond to 8-, 6-, 4-, 3-, and 2-bit quantized models in each line. The model size and GLUE score of 32- and 3-bit BERT-Base and BERT-Large are shown in the figure. }
\label{fig1}
\end{figure*}

We use Arithmetic Computation Effort (ACE)\cite{zhang2021pokebnn}, a newly proposed hardware- and energy-inspired cost metric to evaluate the inference cost of quantized BERT models. ACE is defined as
\begin{equation}
\label{ace}
 ACE=\sum_{i\in I, j \in J} n_{i,j} \cdot i \cdot j
\end{equation}
where $n_{i,j}$ is the number of Multiply-ACcumulate operations (MAC) between a $i$-bit number and a $j$-bit number. $I$ and $J$ are sets of all quantization bits used for inference. ACE is shown to be well correlated to the actual energy consumption on Google TPUs hardware and used to evaluate the inference cost of Binary Neural Networks in \citet{zhang2021pokebnn}.

\subsection{Experimental Results on GLUE}

The performance of 32-bit BERT-Base and BERT-Large and their 8-bit quantized counterparts are listed in Table \ref{glue_benchmark}. ACIQ and OCS outperform LQ on most datasets, resulting in higher average scores. This is consistent with the methodology analysis that ACIQ and OCS further reduce the weights' quantization error by clipping and splitting. Moreover, the 8-bit BERT-Large merely sees a maximum of 1.0 point decrease in the average score. The 8-bit BERT-Base obtained by ACIQ or OCS achieves a higher average GLUE score than the 32-bit baseline. With the backends supporting the 8-bit operation, we are optimistic that quantized models with decent performance and practically faster inference can be attained.

We further quantize BERT-Base and BERT-Large to 6, 4, 3, and 2 bits to explore how their performance deteriorates in Fig. \ref{fig1}, where the $x$ axis denotes the model size corresponding to these quantization bits. As the quantization bit goes lower and lower, OCS gradually becomes dominant with less GLUE score reduction. Eventually, we can quantize BERT-Base and BERT-Large to 3-bit with only a drop of 1.5 and 3.5 points, respectively. Accordingly, the quantized models are about one-tenth the size of 32-bit baselines. When the quantization bit is 2, all models quantized by three methods barely work on the GLUE benchmark.

\begin{figure*}[!th]
\centering
\includegraphics[width=\linewidth]{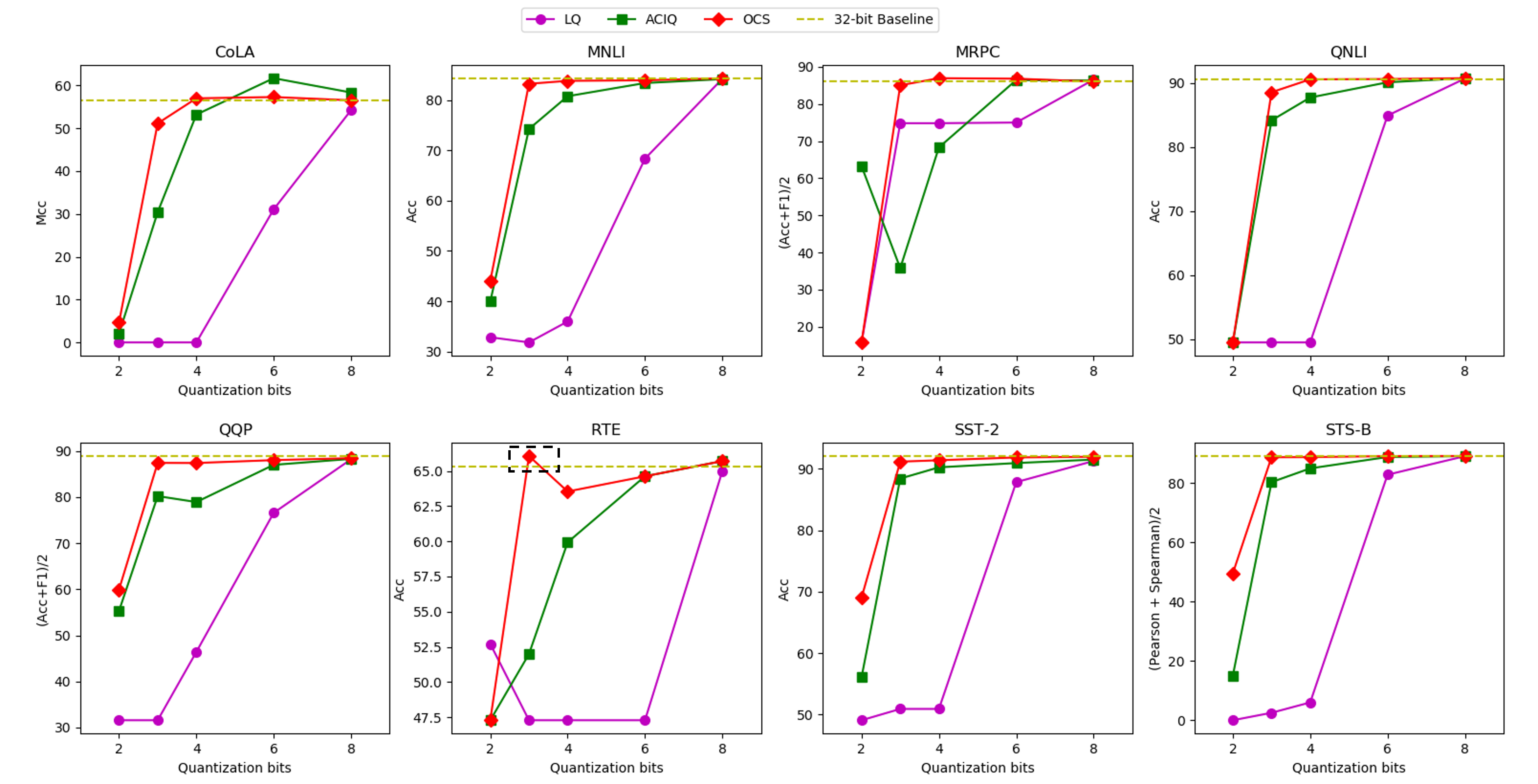}
\caption{The performance of quantized BERT-Base on GLUE tasks. We fine-tune BERT-Base on each task to obtain 32-bit baselines. Then each task-specific baseline is quantized to 8-, 6-, 4-, 3-, and 2-bits. The dotted box in the RTE denotes the case that quantized BERT-base outperforms the 32-bit baseline.}
\label{fig2}
\end{figure*}

\begin{figure*}[!th]
\centering
\includegraphics[width=\linewidth]{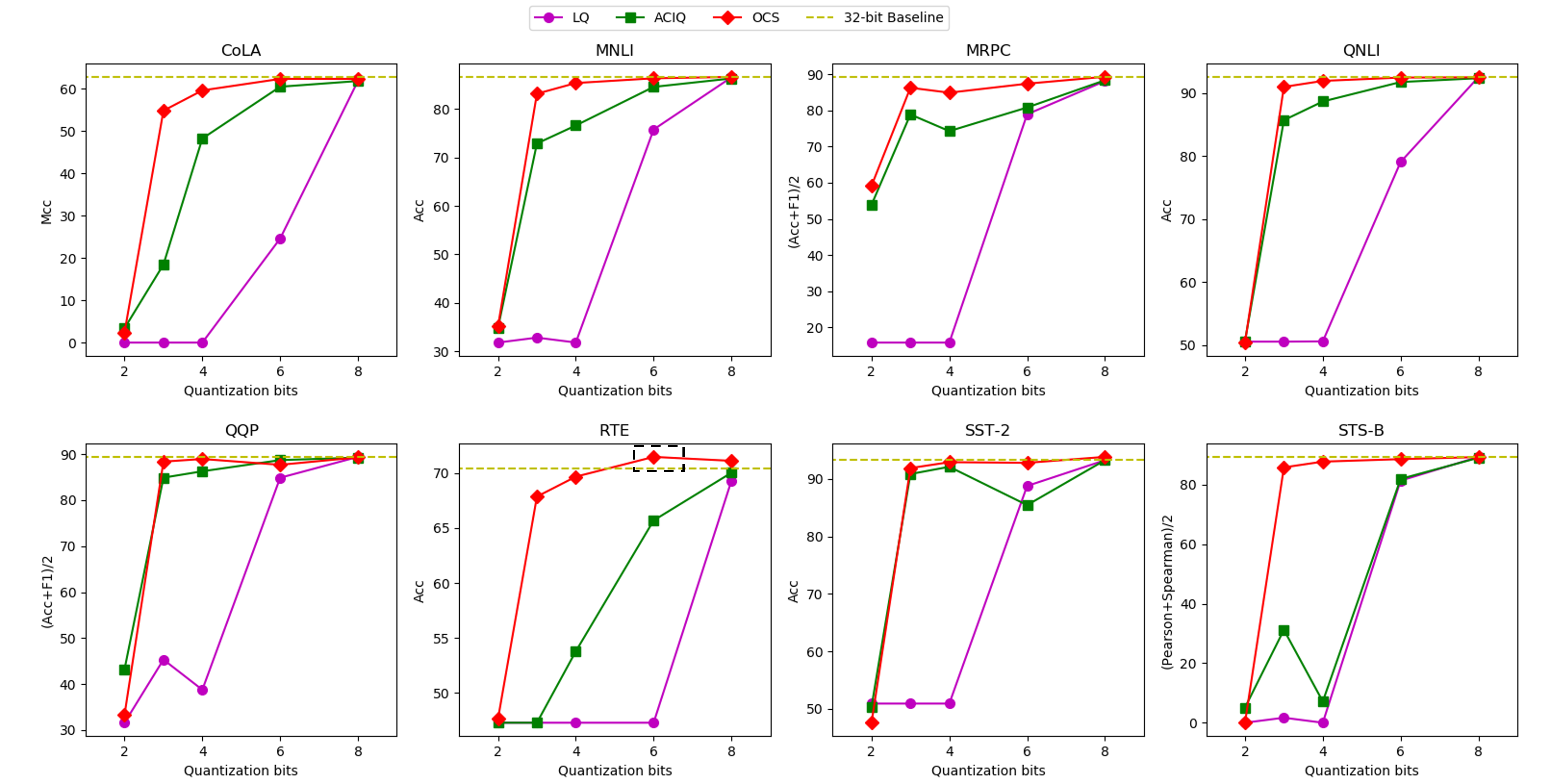}
\caption{The performance of quantized BERT-Large on GLUE tasks. We fine-tune BERT-Large on each task to obtain 32-bit baselines. Then each task-specific baseline is quantized to 8-, 6-, 4-, 3-, and 2-bits. The dotted box represents the case that quantized BERT-Large outperforms the 32-bit baseline on the RTE dataset.}
\label{fig3}
\end{figure*}

Fig. \ref{fig2} and Fig. \ref{fig3} illustrate the performance of quantized BERT-Base and BERT-Large on each downstream task, respectively. LQ incurs the severest performance degradation on all tasks. OCS generally surpasses ACIQ with one exception, which is the 6-bit BERT-base on the CoLA in Fig. \ref{fig2}. On the other hand, BERT-Base quantized by the OCS method outperforms the 32-bit baseline on CoLA, MRPC, and RTE. The quantized BERT-Large surpasses the 32-bit baseline on RTE. These indicate that quantization is more beneficial for small data sets like RTE, where over-parameterization is more likely to occur.

\begin{figure*}[!th]
\centering
\includegraphics[width=0.95\linewidth]{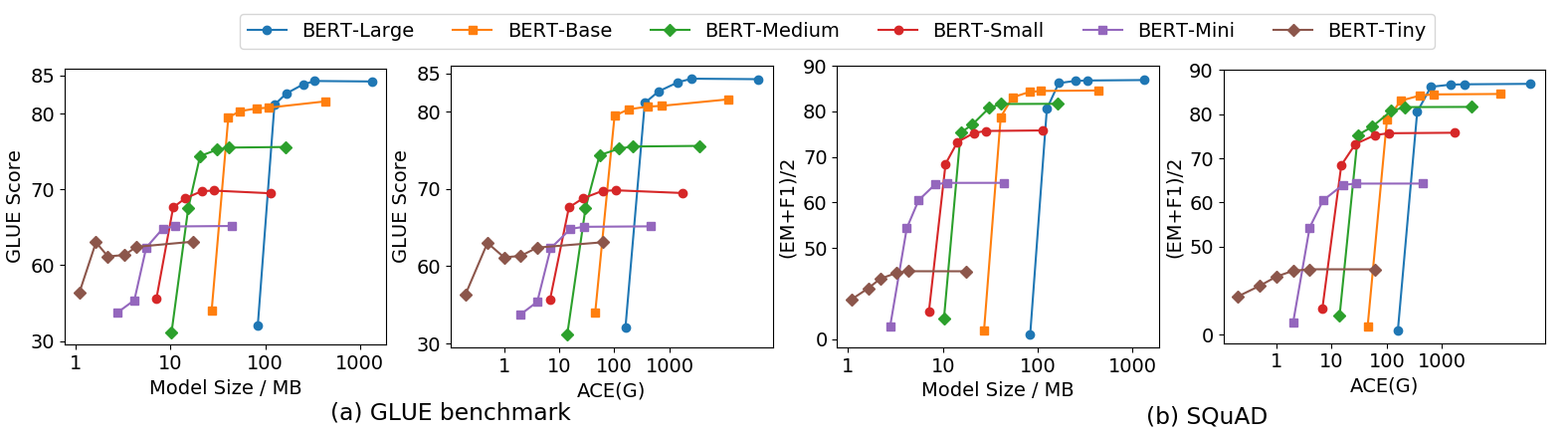}
\caption{The evaluation results of quantized BERT family on (a) GLUE benchmark and (b) SQuAD. The OCS method is employed for quantization here. The points on each line represent 32-, 8-, 6-, 4-, 3-, and 2-bit models from right to left. Arithmetic Computation Effort (ACE) is used to measure the inference cost of quantized models.
} \label{fig4}
\end{figure*}

Considering OCS generally surpasses the other two methods, we further apply OCS to quantize the BERT family, including BERT-Tiny, BERT-Mini, BERT-Small, BERT-Medium, BERT-Base, and BERT-Large, in ascending order of model size. Fig. \ref{fig4} (a) displays their performance versus model size and ACE. Without quantization, a larger-size BERT consistently outperforms the smaller ones, given that more parameters indicate more robust representation capability. However,  Fig. \ref{fig4} (a) implies the possibility that a BERT model with fewer parameters exceeds that with more parameters. For example, 4- and 3-bit BERT-Small surpass 3-bit BERT-Medium with higher GLUE scores, smaller model sizes and less ACE, indicating less memory footprint and inference cost.

Meanwhile, we find that a 3- or 4-bit BERT model typically prevails over the adjacent smaller-sized 32-bit model in the BERT family. For instance, 3-bit BERT-Base outperforms 32-bit BERT-Medium, and 4-bit BERT medium surpasses 32-bit BERT-Small. Quantization brings about more alternatives in the BERT family for environments with limited computational resources.

\subsection{Experimental Results on on SQuAD}

The SQuAD evaluation results of 8-bit BERT-base and BERT-large and their corresponding 32-bit baselines are shown in Table \ref{squad}. Unlike the GLUE benchmark results, 8-bit models do not perform better than 32-bit baselines, while the scores drop by a minimum of 0.1 via the OCS method and a maximum of 0.6 via the LQ method.

\begin{table}[!th]
\small
\begin{center}
    \begin{tabular}{l|cc}
    \hline
     \multicolumn{3}{c}{BERT-base}  \\
    \hline
    & EM & F1  \\
    \hline
    32-bit & \textbf{80.99} & \textbf{88.21}  \\
    LQ & 80.56(-0.43) & 87.93(-0.28)\\
    ACIQ & 80.55(-0.44) & 87.93(-0.28) \\
    OCS & 80.83(-0.16) & 88.17(-0.04)\\
    \hline
     \multicolumn{3}{c}{BERT-large} \\
    \hline 
    & EM & F1 \\
    \hline
    32-bit  & \textbf{83.55} & \textbf{90.24} \\
    LQ &  82.89(-0.66) & 89.79(-0.45)  \\
    ACIQ &  83.25(-0.30) & 90.08(-0.15) \\
    OCS &  83.43(-0.12) & 90.15(-0.09) \\
    \hline
    \end{tabular}
\end{center}
\caption{\label{squad} Results of 8-bit BERT-base and BERT-large on the dev set of SQUAD task. We record the score difference between the baseline and the quantized model in parentheses. All three methods incur a minor performance degradation.}
\end{table}

%\begin{table}[!th]
%\begin{center}
%    \begin{tabular}{l|cc}
%    \hline
%    & BERT-base & BERT-large \\
%    \hline
%    32-bit & \textbf{84.60} & \textbf{86.90} \\
%    LQ & 84.25(-0.35) & 86.34(-0.56) \\
%    ACIQ & 84.24(-0.36) & 86.67(-0.23) \\
%    OCS & 84.50(-0.10) & 86.79(-0.11)\\
%    \hline
%    \end{tabular}
%\end{center}
%\caption{\label{squad} Evaluation results of 8-bit BERT-base and BERT-large on SQUAD. }
%\end{table}

\begin{figure*}[!th]
\centering
\includegraphics[width=0.7\linewidth]{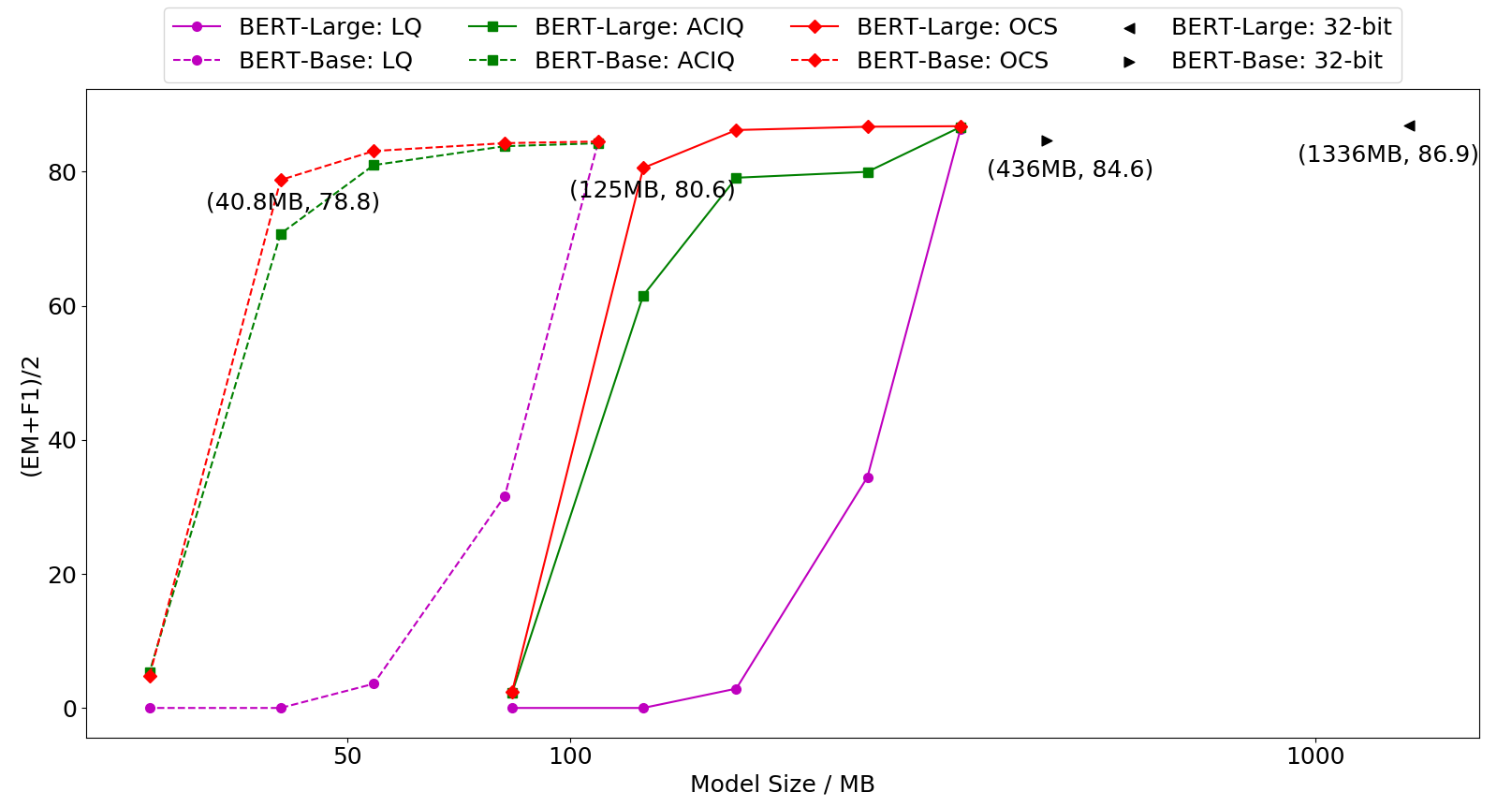}
\caption{The performance of quantized BERT-Base and BERT-Large on SQuAD. Three quantization methods, LQ, ACIQ, and OCS, are applied. The score reported is the average of EM and F1. The points shown from right to left correspond to 8-, 6-, 4-, 3-, and 2-bit quantized models in each line. The model size and the average of EM and F1 score of 32- and 3-bit BERT-Base and BERT-Large are shown in the figure.} \label{fig5}
\end{figure*}

We then quantize BERT-Base and BERT-Large to 8-, 6-, 4-, 3-, and 2-bits and show their performance on SQuAD in Fig. \ref{fig5}. In the end, OCS could quantize BERT-base and BERT-large to 3-bit and retain 93\% of the performance. This portion is lower than that on the GLUE score, indicating that the small datasets on the GLUE benchmark largely contribute to the GLUE Score retainment and could benefit more from quantization.

Furthermore, we quantize the BERT family to different bits via OCS in Fig. \ref{fig4} (b). Apparently, 4-bit BERT-Medium surpasses 32-bit BERT-small, and 3-bit BERT-Small exceeds 32-bit BERT-Mini. A similar phenomenon occurs between every two BERT models in the BERT family. These demonstrate that quantizing a BERT model with more parameters could be more beneficial than directly using a 32-bit model with fewer parameters considering model size and performance. We hope these results could offer some guidance for the distribution of the BERT family in various computation environments.

\section{Conclusions}

In this work, we have quantized fine-tuned BERT-Base and BERT-Large on downstream language tasks: the GLUE benchmark and SQuAD. We apply three post-training quantization methods, including Linear Quantization (LQ), Analytical Clipping Integer Quantization (ACIQ), and Outlier Channel Splitting (OCS), and study how quantized BERT models perform on different tasks. Experimental results show that OCS generally outperforms the other two methods and could quantize BERT-base and BERT-large to 3-bits with a reasonable performance drop. Moreover, a comprehensive evaluation of a set of quantized BERT models with different configurations, ranging from BERT-Tiny to BERT-Large, is presented. According to this, quantizing a BERT model with more parameters could be a better option than simply using a 32-bit model with fewer parameters on resource-constrained devices, considering model size, memory footprint, and performance. In other words, quantization brings about more alternatives for the BERT family by taking the number of bits into account. We hope these could provide a good reference for the practical deployment of the BERT family under resource-limited circumstances. Regarding future work, we would like to quantize both the weights and activations and explore how much the performance can be improved and at what cost.

\bibliography{bibliography}
\bibliographystyle{acl_natbib}

\end{document}